\documentclass[10pt,twocolumn,letterpaper]{article}

\usepackage{iccv}
\usepackage{times}
\usepackage{epsfig}
\usepackage{graphicx}
\usepackage{amsmath}
\usepackage{amssymb}
\usepackage{url}
\usepackage{comment}
\usepackage{subcaption} 
\usepackage{amssymb} 
\usepackage{color}
\usepackage{flushend} 
\usepackage{amsmath} 
\usepackage{epstopdf}
\usepackage{multirow}
\usepackage{bm} 
\usepackage{paralist}
 
\usepackage[pagebackref=true,breaklinks=true,letterpaper=true,colorlinks,bookmarks=false]{hyperref}

\usepackage{hyperref}

\newcommand{\tanhmine}{\operatorname{tanh}} 
\newcommand{\cosmine}{\operatorname{cos}}
\newcommand{\sinmine}{\operatorname{sin}}

\iccvfinalcopy 

\ificcvfinal\fi
\begin{document}
\pagenumbering{gobble} 

\title{View Adaptive Recurrent Neural Networks for High Performance \\ Human Action Recognition from Skeleton Data}

\author{{Pengfei Zhang{\small $~^{1}$}\thanks{This work was done when P. Zhang was an intern at Microsoft Research Asia.}}, ~Cuiling Lan{\small $~^{2}$}\thanks{Corresponding author.}, ~Junliang Xing{\small $~^{3}$}, Wenjun Zeng{\small $~^{2}$}, ~Jianru Xue{\small $~^{1}$}, ~Nanning Zheng{\small $~^{1}$}\\
	\normalsize
	$^{1}$\	Xi'an Jiaotong University, Shannxi, China ~~ $^{2}$\,Microsoft Research Asia, Beijing, China\\
	\normalsize
	$^{3}$\,National Laboratory of Pattern Recognition, Institute of Automation, Chinese Academy of Sciences, Beijing, China \\
	\normalsize
	zpengfei@stu.xjtu.edu.cn,
	\{culan,wezeng\}@microsoft.com, 
	jlxing@nlpr.ia.ac.cn,  
	\{jrxue,nnzheng\}@mail.xjtu.edu.cn
}

\maketitle

\begin{abstract}
	Skeleton-based human action recognition has recently attracted increasing attention due to the popularity of 3D skeleton data. One main challenge lies in the large view variations in captured human actions. We propose a novel view adaptation scheme to automatically regulate observation viewpoints during the occurrence of an action. Rather than re-positioning the skeletons based on a human defined prior criterion, we design a view adaptive recurrent neural network (RNN) with LSTM architecture, which enables the network itself to adapt to the most suitable observation viewpoints from end to end. Extensive experiment analyses show that the proposed view adaptive RNN model strives to (1) transform the skeletons of various views to much more consistent viewpoints and (2) maintain the continuity of the action rather than transforming every frame to the same position with the same body orientation. Our model achieves significant improvement over the state-of-the-art approaches on three benchmark datasets.
\end{abstract}


\section{Introduction}
\label{sec:introduction}
Recognizing human actions has remained one of the most important and challenging problems in computer vision. Demands on human action recognition techniques are growing very fast and have expanded in many domains, such as visual surveillance, human-computer interaction, video indexing/retrieval, video summary, and video understanding \cite{IVC10SurveyAction, CVIU11SurveyAction}.

Considering the differences in inputs, human action recognition can be categorized into color video-based and 3D skeleton-based approaches. While color video based human action recognition has been extensively studied over the past few decades, 3D skeleton based human representation for action recognition has recently attracted a lot of research attention because of its high level representation and robustness to variations of viewpoints, appearances, and surrounding distractions \cite{aggarwal2014human, han2016space, presti20163d, zhang2016rgb}. Biological observations from the early seminal work of Johansson suggest that humans can recognize actions from just the motion of a few joints of the human body, even without appearance information \cite{PP73Perception}. Besides, the prevalence of cost-effective depth cameras such as Microsoft Kinect \cite{zhang2012microsoft}, Intel RealSense \cite{IntelRealSense}, dual camera devices, and the advance of a powerful technique of human pose estimation from depth \cite{CVPR11BestPaper} make 3D skeleton data easily obtainable. Like the many previous works listed in the survey paper \cite{han2016space}, we focus on skeleton-based action recognition.

\begin{figure}[t] 
	\begin{center}
		\includegraphics[width=1\linewidth]{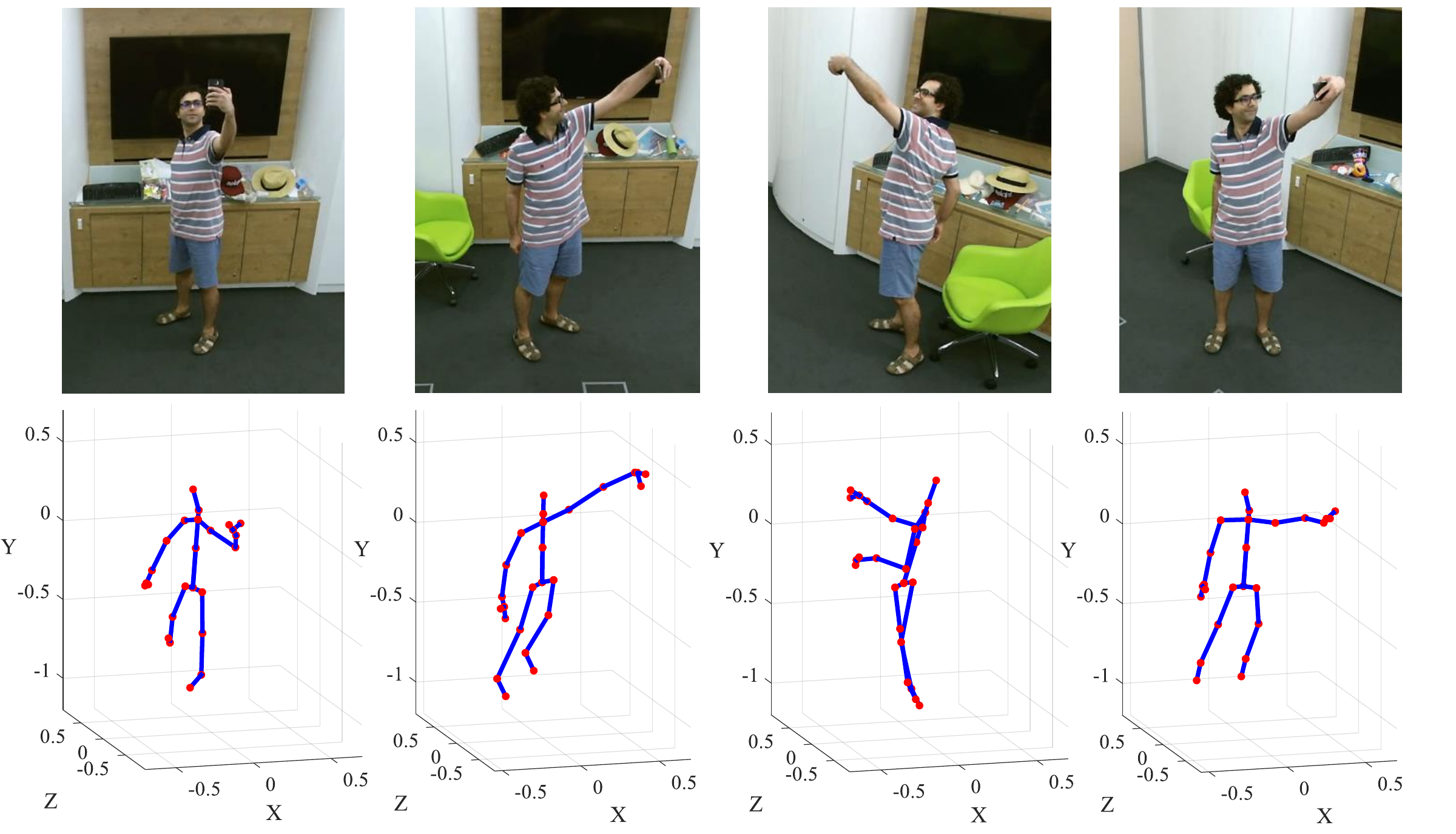}
	\end{center}
	\vspace{-5mm}
	\caption{Skeleton representations of the same posture captured from different viewpoints (different camera position, angle, and the subject orientation) are very different. } 
	\label{fig:Viewpoints}
	\vspace{-3mm}
\end{figure}

One of the main challenges in skeleton-based human action recognition is the complex viewpoint variations when capturing human action data. First, in a practical scenario, the capturing viewpoints of the camera differ among different sequences, \eg, the facing angle, position of the camera, resulting in large differences among skeleton representations. Second, the actor could conduct an action towards different orientations. Moreover, he/she may dynamically change his/her orientations as time goes on. As illustrated in Fig. \ref{fig:Viewpoints}, the skeleton representations of the same posture are rather different when captured from different viewpoints. 
In practice, the variation of the observation viewpoints makes action recognition a very challenging problem \cite{aggarwal2014human, ji2010advances}. Attempts have been made in previous works to overcome the view variations for robust action recognition \cite{ji2010advances, rao2001view, bashir2006feature, shen2008ratio, junejo2008selfsimilarities, Farhadi2008Wrongview, shen2009pointtriplets, weinland2010making, liu2011knowtransfer,iosifidis2012ANN, li2012virtualview, wu2012latentSVM, wu2013cross,mahasseni2013latent,zhang2013virtualpath, rahmani2015knotransfer, feng2015usemoreview}. Most of these works, however, are designed for color video-based human recognition. The investigation of view invariance for skeleton-based human recognition, however, still remains under explored.



\begin{figure*}[htbp] 
	\begin{center}
		\includegraphics[width=0.9\linewidth]{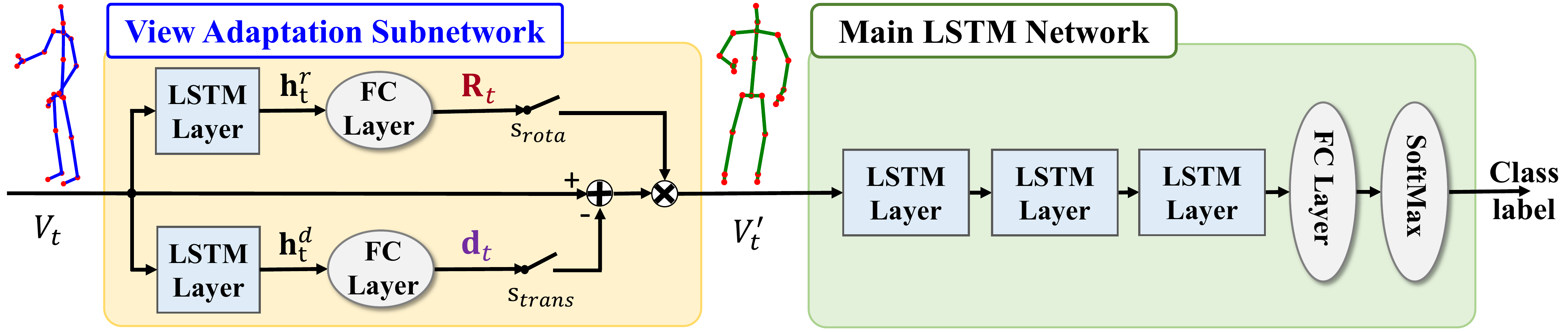}
	\end{center}
	\vspace{-3mm}
	\caption{Architecture of our end-to-end view adaptive RNN, which consists of a View Adaptation Subnetwork, and a Main LSTM Network. The View Adaptation Subnetwork determines the suitable observation viewpoint at each time slot. With the skeleton representations under the new observation viewpoints, the main LSTM network determines the action class.}
	\label{fig:framework}
	\vspace{-2mm}
\end{figure*} 

There are only a few attempts in previous works to consider the effect from view variations. A general treatment employs a pre-processing step to transform the 3D joint coordinates from the camera coordinate system to a person-centric coordinate system by placing the body center at the origin, followed by rotating the skeleton such that the body plane is parallel to the $(x, y)$-plane, to make the skeleton data invariant to absolute location, and the body orientation \cite{CVPR12HO3DJ, vemulapalli2014human, CVPR15HRNN,zhu2015co, jiang2015informative, Shahroudy_2016_CVPR, liu2016spatio,AAAI17Atte}. Such a pre-processing gains partial view-invariant. However, it also has many drawbacks. On one hand, it loses partial motion information, \eg, the moving trajectory and speed of the body center, and the changing dynamics of the body orientation. For example, the action of walking becomes walking in the same place and the action of dancing with body rotating becomes dancing with body facing a fixed orientation. On the other hand, the processing (\ie, translation, rotation) is not explicitly designed with the target of optimizing action recognition in mind but is based on human defined criteria, which reduces the space for exploiting optimal viewpoints. How to design a system which provides superior viewpoint for action recognition is still an under-explored problem, and warrants more investigation.

In this work, we address the view variation problem for high performance skeleton-based action recognition. Instead of processing the 3D skeletons based on human defined criteria for solving view variations, we propose a view adaptation scheme which automatically regulates the observation viewpoint at each frame to obtain the skeleton representation under the new view. Note that the regulation of the viewpoint of the camera is equivalent to the transformation of the skeleton to a new coordinate system. To this end, as shown in Fig.~\ref{fig:framework}, we design a view adaptive RNN with LSTM architecture to learn and determine the appropriate viewpoints based on the input skeleton. The skeleton newly represented in the determined observation viewpoint is used for easier action recognition by a main LSTM network. With the objective of maximizing recognition performance, the entire network is end-to-end trained to encourage the view adaptation subnetwork to learn and determine suitable viewpoints.

To summarize, we make the following contributions.
\vspace{-2mm}
\begin{itemize}
\setlength{\itemsep}{0pt}
\item We propose a self-regulated view adaption scheme which re-positions the observation viewpoints dynamically to facilitate better recognition of the action from skeleton data.
\setlength{\itemsep}{0pt}
\item We integrate the proposed view adaption scheme into an end-to-end LSTM network which automatically determines the ``best" observation viewpoints during recognition.
\setlength{\itemsep}{0pt}
\item We have made many observations and analyses of the results from the view adaptation model. We find that the proposed model automatically regulates the skeletons to more consistent observation viewpoints while maintaining the continuity of an action. 
\end{itemize}
\vspace{-2mm}
Based on the above contributions, we present an end-to-end, high performance action recognition system. Extensive experiment analyses and evaluations demonstrate its strong ability to overcome the view variation problem, and its state-of-the-art performance on three benchmark datasets.

\section{Related Work}
\label{sec:relatedwork}
   
\subsection{View Invariant Action Recognition}
Human actions may be observed from arbitrary camera viewpoints in realistic scenes. This factor is a barrier for the development of efficient action recognition techniques. Researchers have paid much attention to this issue and designed view-invariant approaches for action recognition from color videos \cite{ji2010advances,rao2001view,bashir2006feature,shen2008ratio,junejo2008selfsimilarities, Farhadi2008Wrongview, shen2009pointtriplets, weinland2010making, liu2011knowtransfer,iosifidis2012ANN, li2012virtualview,wu2012latentSVM, wu2013cross,mahasseni2013latent,zhang2013virtualpath, rahmani2015knotransfer,feng2015usemoreview}.
One category of approaches requires multiple view videos for training \cite{iosifidis2012ANN,feng2015usemoreview,weinland2010making,wu2013cross,mahasseni2013latent}. For example, the 3D histogram of Oriented Gradients based Bag of Words model \cite{weinland2010making} is learned from all viewpoints of data to provide robustness to view changes. Another category of approaches designs view-invariant feature representations \cite{junejo2008selfsimilarities,rao2001view,bashir2006feature} like self-similarity descriptors \cite{junejo2008selfsimilarities} or descriptions based on trajectory curvature \cite{rao2001view,bashir2006feature}. There is also a category of approaches that employ knowledge transfer-based models \cite{Farhadi2008Wrongview,liu2011knowtransfer,li2012virtualview,zhang2013virtualpath,zheng2013sparse,rahmani2015knotransfer}. They find a view independent latent space in which features from different views are directly comparable. Considering the different domains of the color videos and skeleton sequences, the approaches designed for color videos cannot be directly extended to skeleton-based action recognition.  

As a comparison, the study of viewpoint influences on skeleton-based action recognition is under-explored. The commonly used strategies are monotonous where a pre-processing of skeleton is performed \cite{CVPR12HO3DJ,vemulapalli2014human,CVPR15HRNN,zhu2015co,jiang2015informative, Shahroudy_2016_CVPR, liu2016spatio,AAAI17Atte}. Unfortunately, they result in the loss of partial relative motion information. Sequence-based pre-processing, which performs the same transformation on all frames with the parameters determined from the first frame so that the motion is invariant to the initial body position and initial orientation, can preserve motion information. However, since the human body is not rigid, the definition of the body plane by the joints of ``hip", ``shoulder", ``neck" is not always suitable for the purpose of orientation alignment \cite {wang2014learning}. After the alignment of such a defined body plane, a person who is bending over will have his/her legs obliquely upward. Wang et al. \cite{wang2014learning} use only the up-right pose frames in a sequence to determine the body plane by averaging the rotation transformation. However, a sequence may not contain an up-right pose.  

In contrast to the above works, we leverage a content-dependent view adaptation model to automatically learn and determine the suitable viewpoints for each frame. 

\subsection{RNN for Skeleton-based Action Recognition}

Earlier works used hand-crafted features for action recognition from the skeleton  \cite{han2016space,CVPR12HO3DJ}. Many recent works leverage the Recurrent Neuron Networks to recognize human actions from raw skeleton input, with feature learning and temporal dynamic modeling achieved by the neuron networks. Du et al. \cite{CVPR15HRNN} proposes an end-to-end hierarchical RNN for action recognition which takes each body part as input to each RNN subnetwork and fuses the output of subnetworks hierarchically. Zhu et al. \cite{zhu2015co} propose the automatic exploration of the co-occurrence of discriminative skeleton joints in an LSTM network using group sparse regularization. In the part aware LSTM model \cite{Shahroudy_2016_CVPR}, the memory unit of the LSTM model is separated to part-based sub-cells to push the network towards learning long-term context representations for each individual part. To learn both the spatial and temporal relationships among joints, the spatial-temporal LSTM network extends the deep LSTM architecture to two concurrent domains, \ie, the temporal domain and the spatial domain \cite{liu2016spatio}. To further exploit joint discriminations, the spatial-temporal attention model \cite{AAAI17Atte} further introduces the attention mechanism into the network to enable it to selectively focus on discriminative joints of the skeleton within one frame, and pay different levels of attention to the outputs from multiple frames.

Most of the above works take the center and orientation aligned skeletons as input to the RNNs, by using the human defined alignment criteria. In contrast, our model automatically determines the observation viewpoints and thus the skeleton representations for efficient action recognition.

\section{RNN and LSTM Overview}
\label{sec:RNN}

\begin{figure}[bh]
	\centering
	\begin{subfigure}[t]{0.3\linewidth}
		\centering\includegraphics[width=\textwidth]{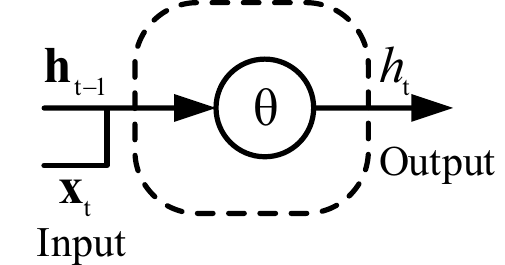}
		\caption{}
		\label{subfig:RNN}
	\end{subfigure}	
	\begin{subfigure}[t]{0.688\linewidth}
		\centering\includegraphics[width=\textwidth]{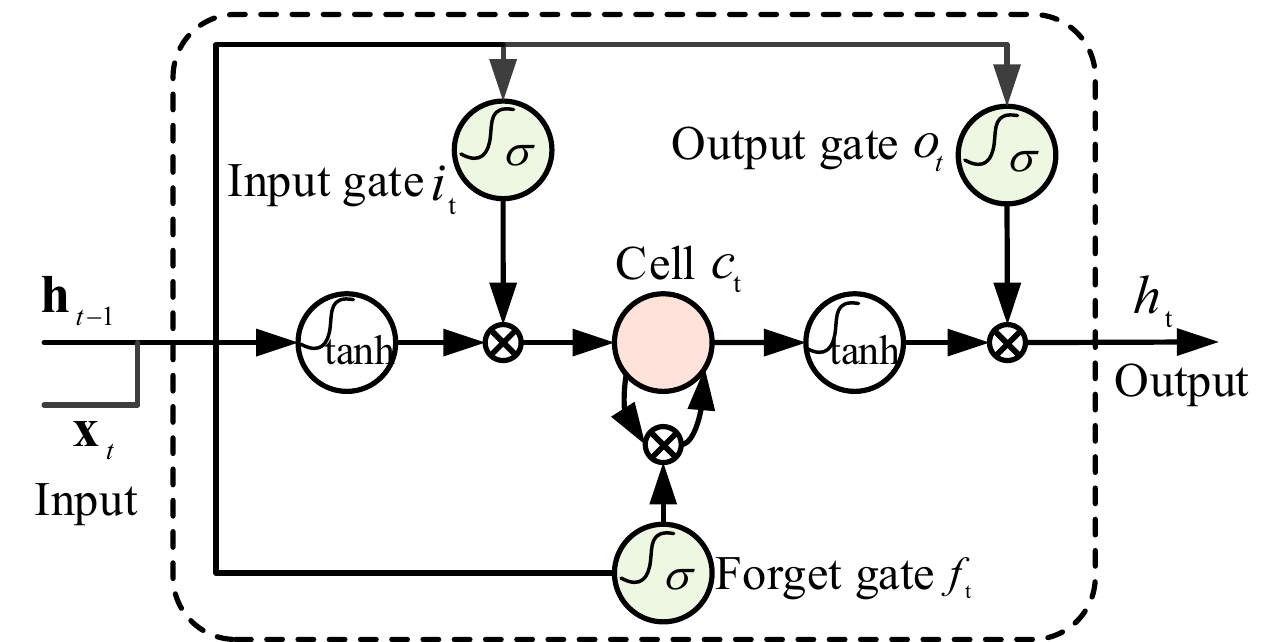}
		\caption{}			
		\label{subfig:LSTM_unit}
	\end{subfigure}
	\vspace{-2mm}
	\caption[]{Structures of the neurons. (a) RNN;
		(b) LSTM.}\label{fig:RNNLSTM}
\end{figure}
To make the paper self-contained, in this section we briefly review the Recurrent Neural Network (RNN), and the RNN with Long Short-Term Memory (LSTM) \cite{LSTM1997}, based on which our framework is built. 

RNN is a powerful model for sequential data modeling and feature extraction, which allows the previous information to persist \cite{Graves2012,LSTM}. Fig.~\ref{fig:RNNLSTM} (a) shows an RNN neuron, where the output response $\mathbf{h}_t$ at time step $t$ is determined by the input $\mathbf{x}_t$ and the hidden outputs from RNN themselves at the last time step $\mathbf{h}_{t-1}$. However, such a standard RNN faces the vanishing gradient effect in practice \cite{LSTM1997,vanish2001,Graves2012}, which is not very capable of handling long-term dependencies. The advanced RNN architecture of LSTM \cite{LSTM1997} mitigates this problem. Fig.~\ref{fig:RNNLSTM} (b) shows an LSTM neuron. The key to LSTM is the cell state $\bf{c}_{t}$, which is kind of like a conveyor belt \cite{LSTM}. The removal of the previous information or addition of the current information to the cell state are regulated with linear interactions by the forget gate $\bf{f}_{t}$ and the input gate $\bf{i}_{t}$.

\section{View Adaptation Model using LSTM}
\label{sec:proposed}
We propose an end-to-end LSTM network with a view adaptation module for skeleton-based human action recognition. Fig.~\ref{fig:framework} shows the overall architecture of the proposed network, which consists of a View Adaptation Subnetwork and a Main LSTM Network. In the following subsections, we first formulate the problem of observation viewpoint regulation. Then we describe our proposed view adaptation network in detail, which is capable of adaptively determining the most suitable observation viewpoints frame by frame.

\subsection{Problem Formulation}
\label{subsec:formulation}
The raw 3D skeletons are recorded corresponding to the camera coordinate system (global coordinate system), with the origin located at the position of the camera sensor. To be insensitive to the initial position of an action and to facilitate our study, for each sequence, we translate the global coordinate system to the body center of the first frame as our new global coordinate system $\mathcal{O}$. Note that the input skeleton $V_{t}$ to our system as in Fig.~\ref{fig:framework} is the skeleton representation under this global coordinate system. 

\begin{figure}[th] 
	\begin{center}
		\includegraphics[width=1\linewidth]{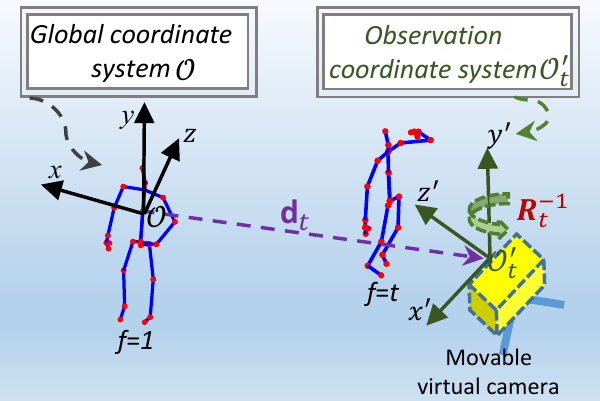}
	\end{center}
	\vspace{-3mm}
	\caption{Illustration of the regulation of the observation viewpoint (movable virtual camera). A skeleton sequence is a record of the skeletons from the first frame $f\!=\!1$ to the last frame $f\!=\!T$ under the global coordinate system $\mathcal{O}$. The action can be re-observed by a movable virtual camera under the observation coordinate systems. For the $t^{th}$ frame, the observation coordinate system is at a new position $\mathbf{d}_t$ with a rotation of $\alpha_t$, $\beta_t$, $\gamma_t$ radians anticlockwise around the $X$-axis, $Y$-axis, and $Z$-axis, respectively, corresponding to the global coordinate system. The skeleton can then be represented under this observation coordinate system $\mathcal{O}'_t$.}
	\label{fig:viewRegulation}
\end{figure}

One can choose to observe an action from suitable views. Thanks to the availability of the 3D skeletons captured from a fixed view, it is possible to set up a movable virtual camera and observe the action from new observation viewpoints as illustrated in Fig.~\ref{fig:viewRegulation}. With the skeleton at frame $t$ re-observed from the movable virtual camera viewpoint (observation viewpoint), the skeleton can be transformed to a representation under the movable virtual camera coordinate system, which is also referred to as the observation coordinate system $\mathcal{O}'_t$. 


Given a skeleton sequence $\mathcal{S}$ with $T$ frames, under the global coordinate system $\mathcal{O}$, the $j^{th}$ skeleton joint on the $t^{th}$ frame is denoted as $\mathbf{v}_{t,j} = [x_{t,j},y_{t,j},z_{t,j}]^{\rm T}$, where $t \in (1,\cdots, T)$, $j \in (1,\cdots, J)$, $J$ denotes the total number of skeleton joints in a frame. We denote the set of joints in the $t^{th}$ frame as $V_t = \{\mathbf{v}_{t,1}, \cdots, \mathbf{v}_{t,J}\}$. 

For the $t^{th}$ frame, assume the movable virtual camera is placed at a suitable viewpoint, with the corresponding observation coordinate system obtained from a translation by $\mathbf{d}_t \in \mathbb{R}^3$, and a rotation of $\alpha_t$, $\beta_t$, $\gamma_t$ radians anticlockwise around the $X$-axis, $Y$-axis, and $Z$-axis,  respectively, of the global coordinate system. Therefore, the representation of the $j^{th}$ skeleton joint $\mathbf{v}'_{t,j} = [x'_{t,j},y'_{t,j},z'_{t,j}]^{\rm T}$ of the $t^{th}$ frame under this observation coordinate system $\mathcal{O}'_t$ is 
\begin{equation}
\label{equ:transform}
\mathbf{v}'_{t,j} = [x'_{t,j},y'_{t,j},z'_{t,j}]^{\rm T} = \mathbf{R}_t \times (\mathbf{v}_{t,j} - \mathbf{d}_t).
\end{equation}
 $\mathbf{R}_t $ can be represented as 
\begin{equation}
\label{equ:Rxyz}
\mathbf{R}_t = \mathbf{R}_{t,\alpha}^x \times \mathbf{R}_{t,\beta}^y \times \mathbf{R}_{t,\gamma}^z,
\end{equation}
where $\mathbf{R}_{t,\gamma}^y$ denotes the coordinate transform for rotating the original coordinate system around the $Y$-axis by $\beta_t$ radians anticlockwise, which is defined as
\begin{equation}
\label{equ:rotation}
\mathbf{R}_{t,\beta}^y = \begin{bmatrix}
\cosmine(\beta_t) & \sinmine(\beta_t) & 0 \\
-\sinmine(\beta_t) & \cosmine(\beta_t) & 0 \\
0 & 0 & 1
\end{bmatrix}.
\end{equation}
Similarly, $\mathbf{R}_{t,\alpha}^x$ and $\mathbf{R}_{t,\gamma}^z$ denote the coordinate transforms for rotating the original coordinate system around the $X$-axis by $\alpha_t$ radians, and around the $Z$-axis by $\gamma_t$ radians anticlockwise, respectively.

Note that all the skeleton joints in the $t^{th}$ frame share the same transform parameters, \ie, $\alpha_t, \beta_t, \gamma_t, \mathbf{d}_t$, considering that the changing of viewpoints is a rigid motion. Given these transform parameters, the skeleton representation $V'_t = \{\mathbf{v}'_{t,1}, \cdots, \mathbf{v}'_{t,J}\}$ under the new observation coordinate can be obtained from (\ref{equ:transform}). Besides, the viewpoints can vary for different frames. The key problem becomes how to determine the viewpoints of the movable virtual camera. 

\subsection{View Adaptive Recurrent Neural Network}
\label{subsec:network}

We use a View Adaptation Subnetwork to automatically determine the observation viewpoints, \ie, $\alpha, \beta, \gamma, \mathbf{d}_t$ (as discussed in section \ref{subsec:formulation}), and use a Main LSTM Network to learn the temporal dynamics and perform the feature abstractions from the view-regulated skeleton data for the action recognition, from end to end, as shown in Fig.~\ref{fig:framework}.

\textbf{View Adaptation Subnetwork.} A regulation of observation viewpoint corresponds to the re-positioning of the movable virtual camera, which can be described by the translation and rotation of this virtual camera (observation coordination system). At a time slot corresponding to the $t^{th}$ frame, with the skeleton $V_{t}$ as input, two branches of LSTM subnetworks are utilized to learn the rotation parameters $\alpha_t, \beta_t, \gamma_t$ to obtain the rotation matrix $\mathbf{R}_t$, and the translation vector $\mathbf{d}_t$, corresponding to the global coordinate system. 

The branch of rotation subnetwork for learning rotation parameters consists of an LSTM layer, and a full connection (FC) layer. The rotation parameters are obtained as
\begin{equation}
	\label{equ:RotationNet}
	[\alpha_t,\beta_t,\gamma_t]^{\rm T} = \mathbf{W}_r \mathbf{h}_t^r + \mathbf{b}_r,
\end{equation}
where $\mathbf{h}_t^r \in \mathbb{R}^{N\times 1}$ is the hidden output vector of the LSTM layer with $N$ denoting the number of LSTM neurons, $\mathbf{W}_r \in \mathbb{R}^{3\times N}$ and $\mathbf{b}_r\in\mathbb{R}^{3\times 1}$ denote the weight matrix and offset vector of the FC layer, respectively. With the rotation parameters, the rotation matrix $\mathbf{R}_t$ is obtained by (\ref{equ:Rxyz}).

The branch of translation subnetwork for learning translation parameters consists of an LSTM layer, and a FC layer. The translation vector $\mathbf{d}_t$ is calculated as
\begin{equation}
\label{equ:TranslationNet}
\mathbf{d}_t = \mathbf{W}_d \mathbf{h}_t^d + \mathbf{b}_d,
\end{equation}
where $\mathbf{h}_t^d \in \mathbb{R}^{N\times 1}$ is the hidden output vector of its LSTM layer, $\mathbf{W}_d \in \mathbb{R}^{3\times N}$ and $\mathbf{b}_d\in\mathbb{R}^{3\times 1}$ denotes the weight matrix and offset vector of the FC layer. Under the observation viewpoint of the $t^{th}$ frame, the representation of the skeleton $V'_t$ is then obtained through (\ref{equ:transform}).

Note that to obtain an efficient view adaptation subnetwork, we have experimented with many alternative designations and found the current design very efficient. First, we use separated LSTM layers for the rotation and translation model learning rather than using shared LSTM layers because the rotation and translation are different operations which are difficult to learn from the shared LSTM neurons. Second, we use the same skeleton input for both the rotation branch subnetwork and the translation branch subnetwork rather than taking the output of one branch (\eg, translation / rotation) as the input of another (\eg, rotation / translation). This is because the learning of the model under the consistent global coordinate system is easier. 
 
\textbf{Main LSTM Network.} The LSTM network is capable of modeling long-term temporal dynamics and automatically learning feature representations. Similar to the designs in \cite{zhu2015co,AAAI17Atte}, we build a main LSTM network by stacking three LSTM layers, followed by one FC layer with a SoftMax classifier. The number of neurons of the FC layer is equal to the number of action classes. 

\textbf{End-to-End Training.} The entire network is end-to-end trainable. We use cross-entropy loss as the training loss \cite{AAAI17Atte}. The gradients of loss flow back not only within each subnetwork, but also from the Main LSTM Network to the View Adaptation Subnetwork. Let us denote the loss back-propagated to the output of the View Adaptation Subnetwork by $\bm{\epsilon}_{v'_{t,j}}$, where $j \in (1,\cdots,J)$ and $J$ is the number of skeleton joints. Then, the loss back-propagated to the output of the branch for determining the translation vector of $\mathbf{d}_t$ is 
\begin{equation}
\label{equ:error_dt}
\bm{\epsilon}_{\mathbf{d}_t} = \sum_{j=1}^{j=J} \frac{\partial \mathbf{v}'_{t,j}}  {\partial \mathbf{d}_t} \odot \bm{\epsilon}_{v'_{t,j}},
\end{equation}
where $\odot$ denotes element-wise product. Similarly, the loss back-propagated to the output of the branch for determining the rotation parameters can be obtained. For example, the loss back-propagated to the output of $\beta_t$ is
\begin{equation}
\label{equ:error_dt}
\bm{\epsilon}_{\beta_t} = \sum_{j=1}^{j=J} \frac{\partial \mathbf{v}'_{t,j}}{\partial \mathbf{R}_t} \frac{\partial \mathbf{R}_t}{\partial \beta_t} \odot \bm{\epsilon}_{v'_{t,j}}.
\end{equation}
With the end-to-end training feasible, the view adaptation model is guided to select the suitable observation viewpoints for enhancing recognition accuracy. 
 
Our scheme has the following characteristics. Firstly, it automatically chooses the suitable observation viewpoints based on the contents, rather than using human predefined criteria. Secondly, the view adaptation model is optimized for the purpose of high accuracy recognition. 

\section{Experiment Results}
We evaluate the effectiveness of our proposed view adaptation scheme on three benchmark datasets. In-depth analyses are made on the NTU dataset. To better understand the model, visualizations of the skeleton representations under the observation viewpoints are given.   

\subsection{Datasets and Settings} 


\textbf{NTU RGB+D Dataset (NTU) \cite{Shahroudy_2016_CVPR}.} This Kinect captured dataset is currently the largest dataset with RGB+D videos and skeleton data for human action recognition, with 56880 video samples. It contains 60 different action classes including daily actions, mutual, and health-related actions. Samples are captured from 17 setups of cameras, where in different setups, the height and distances of the cameras to the subjects are different. For each setup, the three cameras were located at the same height but from different horizontal angles: $-45^o$ (camera 2), $0^o$ (camera 1), $+45^o$ (camera 3). Each subject performed each action twice, once facing towards the left camera and once towards the right camera. Each subject has 25 joints. The standard evaluations include Cross-Subject (CS) evaluation, where the 40 subjects are split into training and testing groups, and Cross-View (CV) evaluation, where the samples of cameras 2 and 3 are used for training while those of camera 1 for testing.


\textbf{SBU Kinect Interaction Dataset (SBU) \cite{yun2012two}.} This Kinect captured dataset is an interaction dataset with two subjects,  containing 282 sequences of 8 classes with subject independent 5-fold cross validation. Each subject has 15 joints. 

\textbf{SYSU 3D Human-Object Interaction Set (SYSU) \cite{hu2015jointly}.} This Kinect captured dataset contains 12 actions performed by 40 subjects. It has 480 sequences. Each subject has 20 joints. We evaluate performance on two standard protocols \cite{hu2015jointly}. For setting-1, half of samples are used for training and the rest for testing for each activity. For setting-2, half of subjects are used for training and the rest for testing. 30-fold cross validation is utilized. Downsampling the sequences in temporal is performed on this dataset in considering that the maximum length of the sequences is high. 

\textbf{Implementation Details.} We build our frameworks based on the platform of Keras \cite{chollet2015keras} toolbox with theano \cite{Theano}. Dropout \cite{srivastava2014dropout} with a probability of 0.5 is used to alleviate overfitting. Gradient clipping similar to \cite{sutskever2014sequence} is used by enforcing a hard constraint on the norm of the gradient (to not exceed 1) to avoid the  exploding gradient problem. Adam \cite{kingma2014adam} is adapted to train all the networks, and the initial learning rate is set as 0.005. 

In our network design, we use 100 LSTM neurons in each LSTM layer for the NTU and the SYSU datasets. To avoid overfitting, we use 50 LSTM neurons in each LSTM layer for the SBU dataset, which has much smaller numbers of training samples than that of the NTU and the SYSU datasets. We set the batch sizes for the NTU, SYSU, and SBU dataset to 256, 64, and 8, respectively. For the View Adaptation Subnetwork, we initialize the full connection layer parameters to zeros for efficient training. 




\subsection{Comparisons to Other State-of-the-Art}

We show the performance comparisons of our proposed view adaptation scheme (\emph{VA-LSTM}) with other state-of-the-art approaches in Table \ref{table:NTU}, Table \ref{table:SBU}, and Table \ref{table:SYSU} for the NTU, SBU and SYSU datasets, respectively. We can see that our scheme significantly outperforms the state-of-the-art approaches by about 6\%, 4\%, 1\% in accuracy for the NTU, SBU, SYSU dataset respectively. 

\subsection{Efficiency of the View Adaptation Model}
\label{subsec:efficiency}

To validate the effectiveness of the proposed view adaptation model, we make two sets of comparisons as summarized in Table \ref{table:In-component}. One set of comparisons evaluates the efficiency among the different pre-processing based methods and our proposed scheme. Another set of results evaluates the efficiency of the view adaptation models. 

\emph{\textbf{VA-LSTM}} is our proposed final view adaptation scheme which automatically regulates the observation viewpoints in the network. This is the scheme where both the translation and rotation branches are connected, \ie, the switch $s_{rota}$ and $s_{trans}$ are on as in Fig.~\ref{fig:framework}. \emph{VA-trans-LSTM} is our scheme which only allows the translation of the viewpoint, \ie, the switch $s_{rota}$ is off while $s_{trans}$ is on. In comparison, \emph{S-trans+LSTM} is our baseline scheme without enabling the view adaptation model, \ie, the switch $s_{rota}$ and $s_{trans}$ are both off, where $V'_t = V_t$. Note that the input $V_t$ is the same as that of our view adaptation schemes, where the global coordinate system is moved to the body center of the first frame for the entire sequence to  be insensitive to the initial position (see section \ref{subsec:formulation}). We refer to this pre-processing as sequence level translation, \ie, \emph{S-trans}. \emph{VA-rota-LSTM} is our scheme which only allows the rotation of the viewpoints, \ie, the switch $s_{rota}$ is on while $s_{trans}$ is off.

\begin{table}[t] 
	\fontsize{8pt}{9pt}\selectfont\centering
	\begin{center}
		\caption{Comparisons on the NTU dataset with Cross-Subject and Cross-View settings in accuracy (\%).} 
		\label{table:NTU}
		\vspace{-2mm}
		\begin{tabular}{c|c|c}
			\hline
			Methods & CS  & CV \\
			\hline
			Skeleton Quads \cite{evangelidis2014skeletal}  & 38.6 & 41.4\\ 
			\hline			
			Lie Group \cite{vemulapalli2014human} & 50.1  & 52.8\\ 
			\hline
			Dynamic Skeletons  \cite{hu2015jointly} & 60.2  & 65.2\\
			\hline
			HBRNN-L  \cite{CVPR15HRNN} & 59.1 &  64.0\\
			\hline
			Part-aware LSTM   \cite{Shahroudy_2016_CVPR} & 62.9 & 70.3  \\
			\hline
			ST-LSTM (Tree Traversal) + Trust Gate \cite{liu2016spatio} & 69.2 & 77.7  \\
			\hline
			STA-LSTM \cite{AAAI17Atte} & 73.4 &  81.2 \\ 
			\hline
			\hline
			VA-LSTM & \textbf{79.4} & \textbf{87.6} \\
			\hline
		\end{tabular}
	\end{center}
	\vspace{-3mm}	
\end{table}

\begin{table}[t] 
	\vspace{-1mm}
	\fontsize{8pt}{9pt}\selectfont\centering
	\begin{center}
		\caption{Comparisons on the SBU dataset in accuracy (\%).} 
		\label{table:SBU}
		\vspace{-2mm}
		\begin{tabular}{c|c}
			\hline
			Methods & Acc. (\%) \\
			\hline
			Raw skeleton \cite{yun2012two} & 49.7 \\
			\hline
			Joint feature \cite{yun2012two} & 80.3 \\
			\hline
			Raw skeleton \cite{ji2014interactive} & 79.4 \\
			\hline
			Joint feature \cite{ji2014interactive} & 86.9 \\
			\hline
			HBRNN-L \cite{CVPR15HRNN} & 80.4 \\
			\hline
			Co-occurrence RNN \cite{zhu2015co} & 90.4 \\
			\hline
			STA-LSTM \cite{AAAI17Atte} & 91.5 \\
			\hline
			ST-LSTM (Tree Traversal) + Trust Gate \cite{liu2016spatio} & 93.3 \\			
			\hline
			\hline 
			VA-LSTM & \textbf{97.2} \\
			\hline
		\end{tabular}
	\end{center}
	\vspace{-3mm}
\end{table}


\begin{table}[t] 
	\fontsize{8pt}{9pt}\selectfont\centering
	\begin{center}
		\caption{Comparisons on the SYSU dataset in accuracy (\%).} 
		\label{table:SYSU}
		\vspace{-2mm}
		\begin{tabular}{c|c|c}
			\hline
			Methods & setting-1  & setting-2 \\
			\hline
			LAFF \cite{hu2016real} &-- & 54.2 \\
			\hline
			Dynamic Skeletons \cite{hu2015jointly} & 75.5 & 76.9 \\			
			\hline
			\hline
			VA-LSTM  & \textbf{76.9} & \textbf{77.5} \\
			\hline
		\end{tabular}
	\end{center}
	\vspace{-6mm}	
\end{table}

From Table \ref{table:In-component}, we observe that the proposed final view adaptation scheme outperforms the baseline scheme \emph{S-trans+LSTM} by 3.4\% and 5.3\% in accuracy for CS and CV settings, respectively, thanks to the introduction of the proposed view adaptation module.

One may wonder how the performance is when using the pre-processed skeletons, basing on the widely used human defined processing criteria, before inputing to the Main LSTM Network. Such pre-processings can be considered as the human defined rules for determining the viewpoints. We name the pre-processing based schemes in the manner of \emph{C+LSTM}, where \emph{C} indicates the pre-processing strategy, \eg, \emph{F-trans+LSTM}. The $3^{rd}$ to  $7^{th}$ rows show the performance of schemes using different pre-processing strategies. \emph{F-trans} means performing frame level translation to have the body center at the coordinate system origin for each frame. \emph{S-rota} means the sequence level rotation with the rotation parameters calculated from the first frame, which is to fix the $X$-axis to be parallel to the vector from ``left shoulder" to ``right shoulder", $Y$-axis to be parallel to the vector from ``spline base" to ``spine", and $Z$-axis as the new $X\!\times\!Y$. Similarly, \emph{F-rota} means the frame level rotation. \emph{F-trans}$\&$\emph{F-rota} means both \emph{F-trans} and \emph{F-rota} are performed, which is similar to the pre-processing in \cite{Shahroudy_2016_CVPR, liu2016spatio,AAAI17Atte}. The scheme \emph{Raw+LSTM}  in the $2^{nd}$ row denotes a scheme which uses the original skeleton without any pre-processing as the input to the Main LSTM Network. Note that for 3D skeletons, the distance of a subject to the camera does not influence the scale of the skeletons. Therefore, the scaling operation is not considered in our framework. 
\begin{table}[t] 
	\centering
	\fontsize{8pt}{9pt}\selectfont\centering
	\caption{Comparisons of pre-processing methods and our view adaptation model on the NTU dataset in accuracy (\%).}
	\label{table:In-component}%
	\vspace{-1mm}
	\begin{tabular}{c|c|c|c}
		\hline
		& Methods & CS & CV \\
		\hline
		wo/ pre-proc. & Raw + LSTM & 66.3 & 73.4\\
		\hline
		\multirow{5}[3]{*}{Pre-proc.} & S-trans + LSTM & 76.0 & 82.3\\
		\cline{2-4}       & F-trans + LSTM & 75.1 & 80.5 \\
		\cline{2-4}       & S-trans$\&$S-rota + LSTM & 76.4 & 85.4 \\
		\cline{2-4}       & S-trans$\&$F-rota + LSTM & 75.0 & 85.1 \\
		\cline{2-4}       & F-trans$\&$F-rota + LSTM & 74.1 & 83.9 \\
		\hline
		\multirow{3}[3]{*}{View adap.} & VA-trans-LSTM & 77.7 & 84.9 \\
		\cline{2-4}       & VA-rota-LSTM & 79.4 & 87.1 \\
		\cline{2-4}       & \textbf{VA-LSTM} & \textbf{79.4} & \textbf{87.6} \\
		\hline
	\end{tabular}%
	\vspace{-2mm}
\end{table}%


From the comparisons in Table \ref{table:In-component}, we have the following observations and conclusions. (1) Our final scheme significantly outperforms the commonly used pre-processing strategies. In comparison with  \emph{F-trans}$\&$\emph{F-rota}+\emph{LSTM} \cite{Shahroudy_2016_CVPR, liu2016spatio,AAAI17Atte}, our scheme achieves improvement by 5.3\% and 3.7\% in accuracy for CS and CV settings, respectively. In comparison with \emph{S-trans}$\&$\emph{S-rota}+\emph{LSTM}, our scheme achieves improvement by 3.0\% and 2.2\% in accuracy. (2) When only the rotation (or the translation) is allowed for adjusting the viewpoints, our scheme still consistently outperforms the schemes with human defined rotation (or translation) pre-processing. (3) Frame level pre-processing is inferior to the sequence level pre-processing, because the former loses more information, \eg, the motion across frames. (4) Being insensitive to the initial position of an action, \emph{S-trans+LSTM} significantly outperforms the scheme with raw skeletons as input \emph{Raw+LSTM}. 

\subsection{Visualization of the Learned Views}

\begin{figure}[bh] 
	\begin{center}
		\includegraphics[width=1\linewidth]{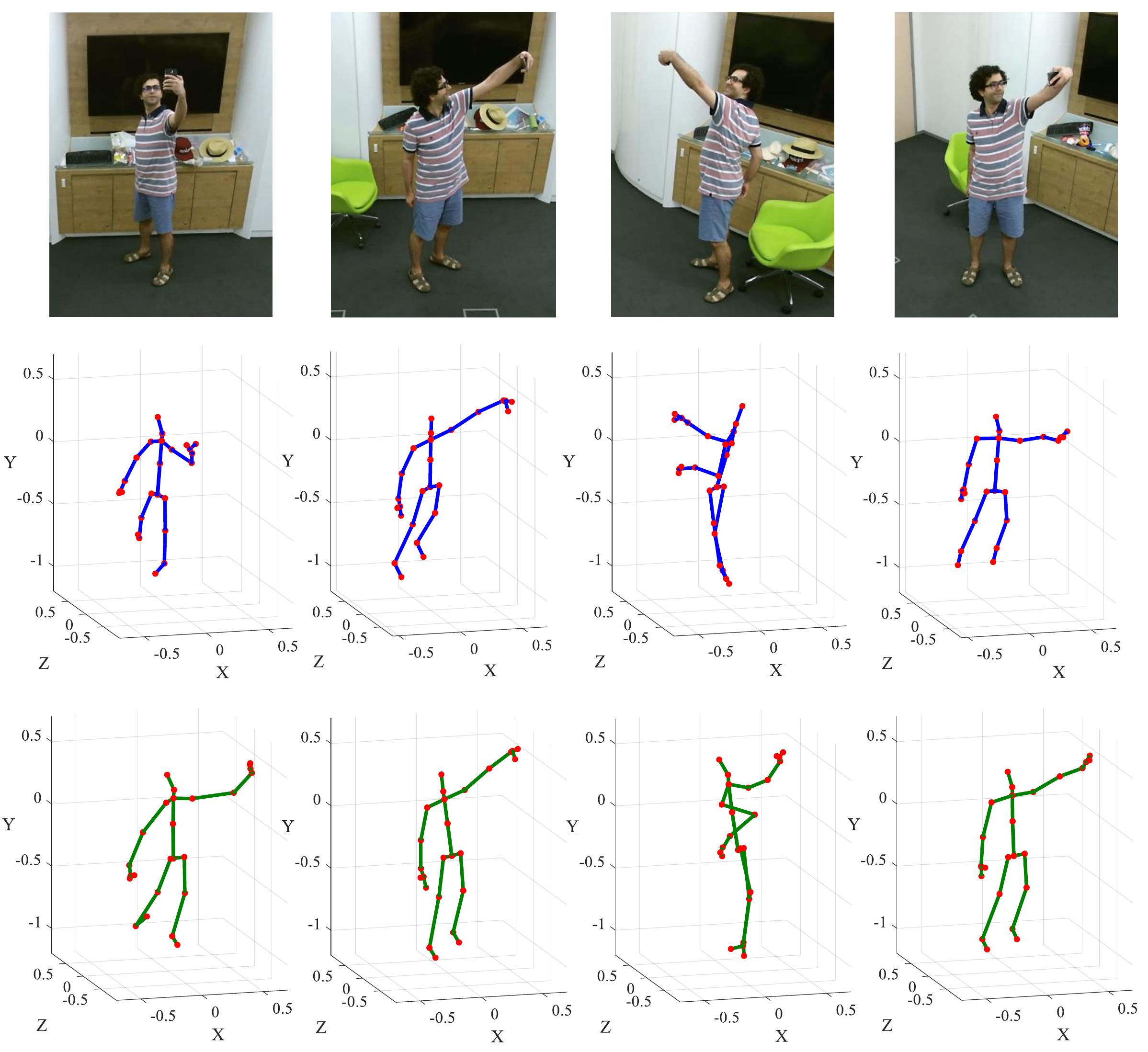}
	\end{center}
	\vspace{-4mm}
	\caption{Frames of the same posture captured from different viewpoints for the same subject. $2^{nd}$ row: original skeletons. $3^{rd}$ row: skeleton representations from the observation viewpoints of our model. Note the third skeleton is very noisy due to occlusion during Kinect shooting.
	} 
	\label{fig:selfie-VA}
\end{figure}
\begin{figure}[bh] 
	\vspace{-1.8mm}
	\begin{center}
		\includegraphics[width=1\linewidth]{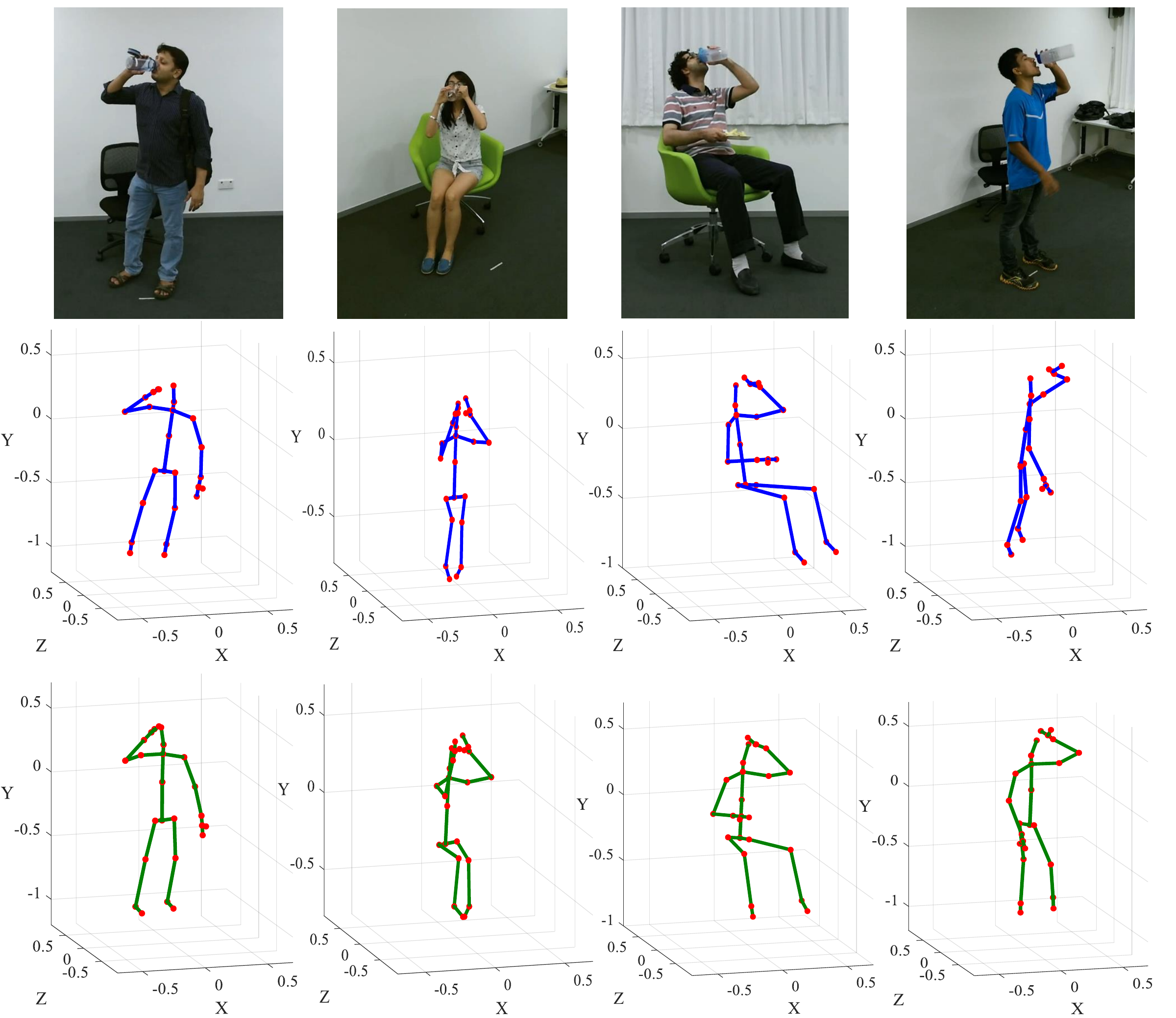}
	\end{center}
	\vspace{-4mm}
	\caption{Frames of the same action ``drinking" captured from different viewpoints for different subjects. $2^{nd}$ row: original skeletons. $3^{rd}$ row: skeleton representations from the observation viewpoints of our model.} 
	\label{fig:Results-actors-VA}
	\vspace{-3mm}
\end{figure}

\begin{figure*}[t] 
	\centering
	\begin{subfigure}[t]{0.48\linewidth}
		\centering\includegraphics[width=\textwidth]{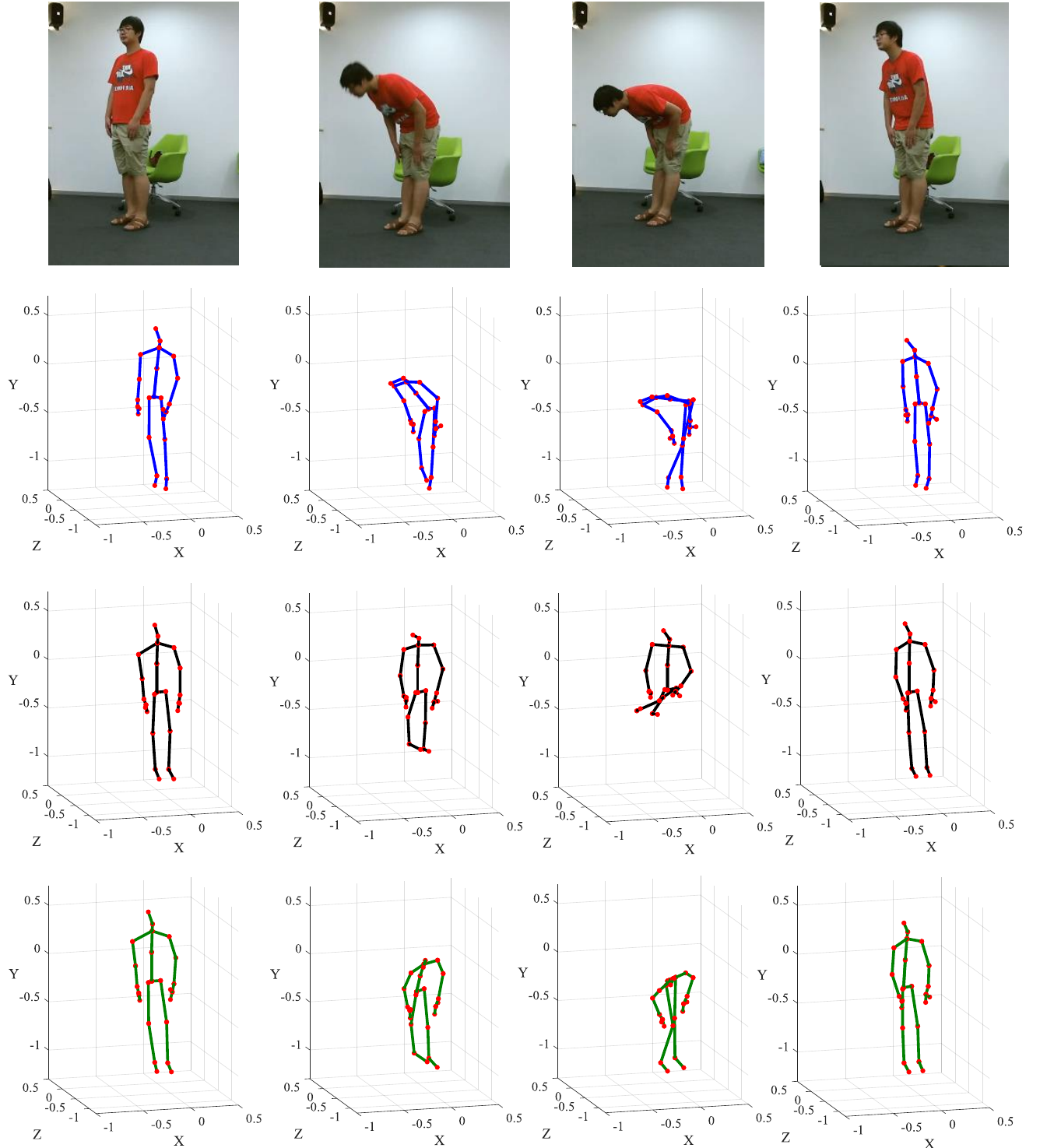}
		\vspace{-5mm}
		\caption{}
		\label{subfig:bow}
	\end{subfigure}	
	~
	\begin{subfigure}[t]{0.48\linewidth}
		\centering\includegraphics[width=\textwidth]{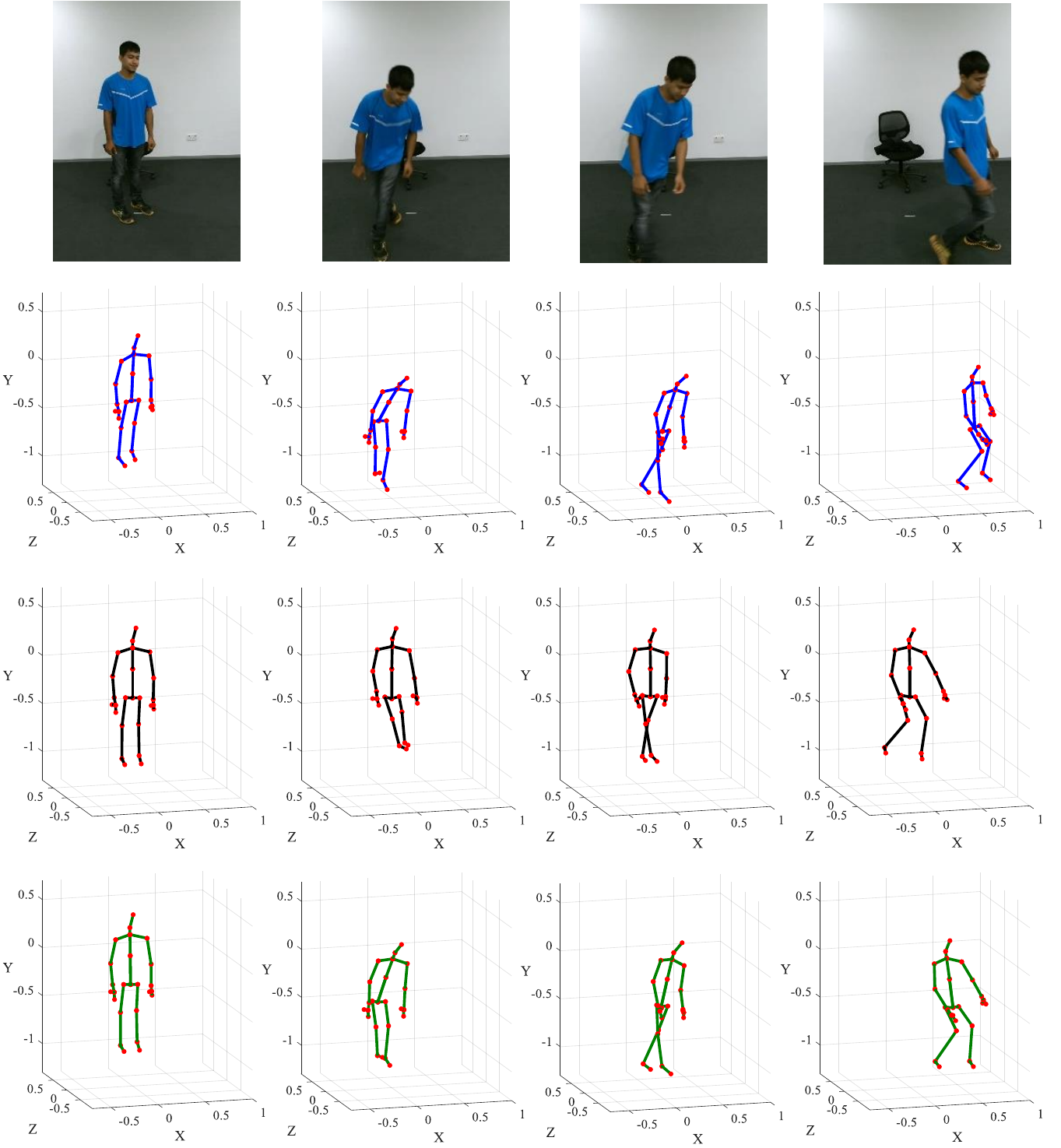}
		\vspace{-5mm}
		\caption{}			
		\label{subfig:staggering}
	\end{subfigure}
	\vspace{-3mm}
	\caption[]{Frames from sequences of actions: (a) ``bow"; (b) ``staggering". $2^{nd}$ row: original skeleton. $3^{rd}$ row: skeleton after the pre-processing with \emph{F-trans}$\&$\emph{F-rota}. $4^{th}$ row: skeleton representation from the observation viewpoints of our model.}\label{fig:action-sequence}
\end{figure*}

At each frame, the view adaptation subnetwork determines the  observation viewpoint (by re-localizing the virtual movable camera) and then transforms the input skeleton $V_t$ to the representation $V'_t$ in the new viewpoint for optimizing recognition performance. We visualize the representations $V_t$ and $V'_t$ for better understanding of our model.

Fig.~\ref{fig:Viewpoints} shows the skeletons from different sequences captured from different viewpoints of the same posture. Interestingly, the transformed skeletons (green) of various viewpoints have much more consistent viewpoints, i.e., frontal viewpoint here. Another example is shown in Fig.~\ref{fig:Results-actors-VA} with the skeleton frames of  the same action performed by different subjects. We can see that they are transformed to similar viewpoints. A similar phenomenon is observed in different actions and sequences.

To visualize the skeleton representations in the sequence along time, we show some frames of an action under the original and new observation viewpoints in Fig.~\ref{fig:action-sequence}. We can see that after our view adaptation model is applied, the subjects even for different actions are oriented toward a more consistent view. Different from frame level pre-processing (as in the $3^{rd}$ row), the transformed skeletons among frames are continuous and looks much natural. In Fig.~\ref{fig:action-sequence}~(a) of action ``bow", the orientation of the body after the processing of our model is parallel to $X$-axis while the legs after frame level pre-processing becomes obliquely upward. In Fig.~\ref{fig:action-sequence}~(b) of action ``staggering", the position changes of the subject after the processing of our model remain while such motion is lost for the pre-processing results. 

From observations, we find that the learned view adaptation model tends to (1) regulate the observation viewpoints to present the subjects as if observed in a consistent viewpoint cross sequences and actions; (2) maintain the continuity of an action without losing much of the relative motions. 

Optimized with the target of maximizing the recognition performance, the proposed view adaptation model is much effective in choosing the suitable viewpoints. The consistency of viewpoints for various actions/subjects overcomes the challenge caused by the diversity of viewpoints in video capturing, enabling the network to focus on the learning of action-specific features. Besides, unlike some pre-processing strategy, the valuable motion information is preserved.    

\section{Conclusion}

We present an end-to-end view adaptation model for human action recognition from skeleton data. Instead of following the human predefined criterion to re-position the skeletons for action recognition, our network is capable of regulating the observation viewpoints to the suitable ones by itself, with the optimization target of maximizing recognition performance. It overcomes the limitations of the human defined pre-processing approaches by exploiting the optimal viewpoints through the content dependent recurrent neuron network model. Experiment results demonstrate that the proposed model can significantly improve the recognition performance on three benchmark datasets and achieve state-of-the-art results. 

\section*{Acknowledgements}
Junliang Xing is partly supported by the Natural Science Foundation of China (Grant No. 61672519), Jianru Xue is partly supported by National Key Research and
Development Plan 2016YFB1001004.

{\small
\bibliographystyle{ieee}
\bibliography{egbib}
}

\end{document}